\documentclass[10pt,a4paper,oneside,onecolumn]{article}

\usepackage{TRA}
\usepackage{enumerate}

\usepackage{amsmath,amssymb}
\usepackage{graphicx}
\usepackage{hyperref}
\usepackage{caption}
\usepackage{subcaption}
\usepackage{algorithm}
\usepackage{algpseudocode}
\usepackage{tikz}
\usepackage{easyReview}
\usetikzlibrary{shapes.geometric, arrows, calc}
\usepackage{multirow}
\usepackage{easyReview}
\usepackage{acronym}
\usepackage[style=numeric, backend=biber, sorting=none]{biblatex}

\newacro{ODD}{Operational Design Domain}
\newacro{AI}{Artificial Intelligence}
\newacro{xAI}{Explainable AI}
\newacro{SIL}{Safety Integrity Level}

\newtheorem{definition}{Definition}

\title{Building Trust in Artificial Intelligence: A Necessity for Railway Applications}
\author{ Lefebvre Renard Clément\textsuperscript{1*[0009-0006-3917-6902]}, Lébé
        Vincent\textsuperscript{2[0009-0005-2114-1261]}, \\
        Da Silva Ribeiro Pereira Ricardo\textsuperscript{1[0009-0006-8969-2753]},
        Sundell Johan\textsuperscript{3[0000-0001-7212-5644]}, Jaoul
        Arnaud\textsuperscript{2[0009-0006-3473-7491]}, \\
        Saiah Kenza\textsuperscript{2[0009-0008-6131-8780]}, Mijatović
        Nenad\textsuperscript{4[0000-0003-0071-9683]}\\
} \affiliation{ \textsuperscript{1}Alstom - Mobility Data Science and AI, Madrid,
Spain\\
         \textsuperscript{2}Alstom - Mobility Data Science and AI, Saint-Ouen, France\\
         \textsuperscript{3}Alstom - Safety Engineering, Stockholm, Sweden\\
         \textsuperscript{4}Alstom - Chief AI \& Data Science Office, Pittsburgh,
         Pennsylvania USA\\
\noindent
E-mails:
\href{mailto:clement.lefebvre-renard@alstomgroup.com}
{*clement.lefebvre-renard@alstomgroup.com},
\href{mailto:vincent.lebe@alstomgroup.com}
{vincent.lebe@alstomgroup.com},
\href{mailto:clement.lefebvre-renard@alstomgroup.com}
{ricardo.pereira@alstomgroup.com},
\href{mailto:ricardo.pereira@alstomgroup.com}
{johan.sundell@alstomgroup.com},
\href{mailto:johan.sundell@alstomgroup.com}
{arnaud.jaoul@alstomgroup.com},
\href{mailto:kenza.saiah@alstomgroup.com}
{kenza.saiah@alstomgroup.com},
\href{mailto:nenad.mijatovic@alstomgroup.com}{nenad.mijatovic@alstomgroup.com}
}
\begin{document}
\maketitle

\begin{abstract}
\ac{AI} is currently only applied to non-safety critical applications due to the strict
standards and regulations for railway industries. We propose to review the three main fields necessary to increase trust in data
science and \ac{AI} algorithms and reach compliance: robustness, \ac{ODD}, and
explainability. Robustness is the ability of an \ac{AI} system to maintain its level of
performance under any circumstances (ISO24029). \ac{ODD}s
allow the explicit definition of operating conditions under which a system is intended
to operate, according to the recently published DIN DKE SPEC 99004. Explainability is the property of an \ac{AI} system to express
important factors influencing the \ac{AI} system results in a way that humans can
understand. Those 3 domains of research are already well
investigated by non-railway actors, with algorithms and methods ready to use for railway
applications. A system view is necessary to ensure all trustworthy requirements interact
continuously in a safe MLOps environment thereby fostering acceptance from regulators,
operators and the public. Beyond safeguarding safety-critical applications, we aim to
show that fostering deep trust in \ac{AI}, as now required by regulatory frameworks
worldwide, will unlock its full potential and transform the pace of adoption across
mission-critical domains.\end{abstract}

\keywords{ Artificial Intelligence, Railway, Transportation, Mobility, Trust,
Safety. }

\section{Introduction}

The levers of trust needed for \ac{AI}, as described by Bo Li et al.
\cite{Li2023TrustworthyAI} (\citeyear{Li2023TrustworthyAI}), are: Robustness,
Explainability, Transparency, Reproducibility, Generalization, Fairness, Privacy and
Accountability. The theoretical framework they propose towards a more Trustworthy
\ac{AI} is very complete and valuable, as models are increasingly impacting our way of
working. This work addresses the industrial gaps of the railway industry to deploy
\ac{AI} at scale, including in safety-related projects. Our paper focuses on the three
domains where research is necessary to unlock deploying \ac{AI} when trust is paramount:
Robustness, Explainability, \acf{ODD}. This publication uses the terminology presented
in the ISO22989 \cite{ISO22989} (2022), where ”AI system” is defined in this standard as an
engineered system that generates outputs such as content, forecasts, recommendations or
decisions for a given set of human-defined objectives.

\subsection{Regulatory Landscape}

The regulatory environment for \ac{AI} in the railway sector is evolving rapidly, and is
shaped by both sector-specific safety standards and cross-sectoral \ac{AI} regulations.
A key development is the EU \ac{AI} Act, defined as Regulation (EU) 2024/1689
\cite{eu-ai-act-2024}, which introduces a risk-based classification of \ac{AI} systems
into the following four categories: Unacceptable risk (e.g., social scoring), High risk
(e.g., \ac{AI} in safety functions or critical infrastructure), Specific transparency
obligations (e.g., bots, deepfakes), Minimal or no risk. Given the increasing use of
\ac{AI} in predictive maintenance, autonomous train operations, and safety-critical
systems, many railway applications fall under the high-risk category. This
classification imposes stringent requirements for transparency, traceability, and human
oversight, making \ac{xAI} not only a compliance necessity but also a valuable tool to
better understand subsystems. To support these regulatory demands, several international
standards and technical reports provide foundational guidance:
\begin{itemize}
    \item ISO/IEC TR5469:2024 \cite{ISOIEC_TR_5469_2024}
    (\citeyear{ISOIEC_TR_5469_2024}) plays a crucial interim role by offering practical
    guidance on the use of \ac{AI} in safety-related systems while the more detailed
    ISO/IEC TS22440 is under development. It addresses the functional safety of \ac{AI}
    systems, particularly in E/E/PE safety-related systems, and introduces a three-stage
    realization principle (data acquisition, knowledge induction, and output
    generation). 
    and mitigation strategies tailored to AI.
    \item IEC61508 \cite{IEC61508-2010} (2010), the cornerstone standard for functional
    safety of electrical / electronic / programmable electronic (E/E/PE) systems, provides the
    conceptual backbone for ISO/IEC TR5469. Many of the safety principles, such as risk
    reduction, ALARP (As Low As Reasonably Practicable), and positive risk balance, are adapted and extended to address the
    unique characteristics of \ac{AI} systems.
    \item ISO/IEC22989 \cite{ISO22989} (2023) emphasizes
    explainability, ensuring \ac{AI} decisions are interpretable by humans.
    \item ISO/IEC24029 \cite{iso24029part1_and_2} (\citeyear{iso24029part1_and_2})
    focuses on robustness, defining it as the \ac{AI} system’s ability to maintain
    performance under varying conditions.
    \item DIN DKE SPEC 99004 \cite{DIN99004} (\citeyear{DIN99004}) introduces the
    concept of Operational Design Domains (ODD), specifying the operational boundaries
    for \ac{AI} systems.
\end{itemize}
ISO/IEC TR5469 also highlights the limitations of traditional software assurance methods
when applied to AI, such as the inapplicability of code coverage metrics and the
challenge of verifying non-deterministic, data-driven models. It advocates for
architectural safeguards, robust learning techniques, and runtime monitoring to mitigate
AI-specific risks. In summary, legal frameworks like the EU \ac{AI} Act provide
compliance obligations and technical standards provide methodological tools. Those are
not sufficient to state clear guidelines for AI compliance in railway safety critical
applications, hence a growing pressure from all the stakeholders to clarify the
requirements.

\subsection{Safety Integrity Level}
\ac{SIL} is used to evaluate and specify the safety requirements for critical system
functions within railway operations and introduces a probabilistic safety approach. The
\ac{SIL} rating system appears in several standards applicable on railway solutions, for
instance in IEC61508 \cite{IEC61508-2010} (2010) for electrical, electronic and
programmable electronic (E/E/PE) systems or in EN50126 \cite{EN50126}
(\citeyear{EN50126}), EN50716 \cite{EN50716} (\citeyear{EN50716}), and EN50129
\cite{EN50129} (\citeyear{EN50129}) for safe software development in European Railway
Applications. The \ac{SIL} level and its tolerable hazard rate associated are described
in Table \ref{tab:sil}.

\begin{table}[h!]
\centering
\begin{tabular}{|c|c|}
\hline
Safety Integrity Level & Tolerable Hazard Rate (THR) per hour and per function \\ \hline
4 & $10^{-9} \leq \text{THR} < 10^{-8}$ \\ \hline
3 & $10^{-8} \leq \text{THR} < 10^{-7}$ \\ \hline
2 & $10^{-7} \leq \text{THR} < 10^{-6}$ \\ \hline
1 & $10^{-6} \leq \text{THR} < 10^{-5}$ \\ \hline
\end{tabular}
\caption{Safety Integrated Levels}
\label{tab:sil}
\end{table}

Peter Wigger (\citeyear{Example_SIL}) \cite{Example_SIL} presents several examples of
Railway applications \ac{SIL} levels on the Copenhagen Metro subsystems, where some
functions of the ATP (Automatic Train Protection) such as the interlocking and speed
profile control are SIL4 while the vehicles door management is SIL3. A SIL4 solution
would translate into allowing a failure every 100 000 years, which corresponds to the
highest level of safety.

While \ac{AI} models typically aim for 80–99\% accuracy, this falls short of the
stringent tolerable hazard rates required by \ac{SIL} standards. Even high-performing
models remain vulnerable to previously unseen out-of-distribution data. Current
standards do not fully address AI-specific challenges, especially in safety-critical
domains like railways, making compliance assessment difficult and underscoring the need
for dedicated \ac{AI} safety specifications. In operational settings like maintenance
analysis, users can benefit from understanding \ac{AI} outputs, as proposed by Di-Santi
et al. (\citeyear{PM}) for the predictive maintenance of track circuits \cite{STDS}
\cite{CVCM} and point machines \cite{PM} using neural networks. However, in
safety-critical scenarios such as an obstacle detected on rails, lives are at stake,
leaving no room for errors. In such cases, trust in the system and higher Safety
Integrity Levels (SIL) becomes essential, which is why AI is not used currently. This
publication describes the need for research to design robust \ac{AI} systems embedded in
\ac{SIL} environments.

\section{Robustness of \ac{AI} Systems}

\subsection{Definition and context}

Robustness refers to the ability of an \ac{AI} system to maintain its level of
performance under any circumstances (ISO24029 \cite{iso24029part1_and_2},
\citeyear{iso24029part1_and_2}). This concept addresses the extent to which the system
performs under expected and unseen perturbations in input data due to various
operational environments. In the railway context, this means that system performance
must be maintained across diverse and potentially harsh conditions (such as lighting,
weather, \dots). Collecting diverse and representative data is often difficult due to
infrastructure constraints, privacy concerns, or the rarity of edge-case scenarios. As a
result, models trained on limited data may become highly sensitive to
out-of-distribution inputs.

In the AI literature, generalization and robustness are separate concepts.
Generalization refers to the ability of a model to perform well on unseen data, whereas
robustness is mainly defined as a per-input property, meaning that a model is robust if
it does not change its output for a specified set of perturbations on a given input
\cite{szegedy2014intriguing, goodfellow2015explaining}.

\subsection{Adversarial robustness}

\begin{figure}[ht]
    \centering
    \includegraphics[width=0.4\linewidth]{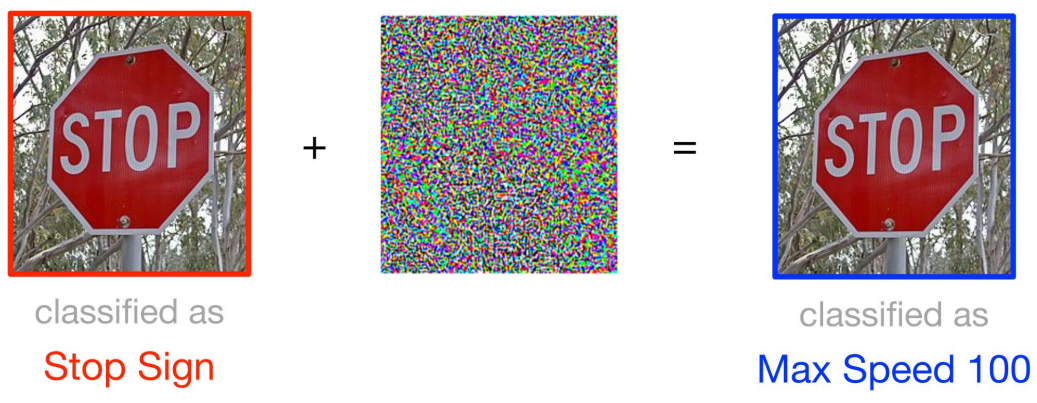}
    \caption{Example of adversarial attack on signaling classification from
    \cite{julia2023data}.}
    \label{fig:adversarial-example}
\end{figure}

In the field of evaluating the robustness of machine learning models, adversarial
examples are gathering more and more interest from the community
\cite{goodfellow2015explaining, guo_comprehensive_2023, szegedy2014intriguing,
carlini_towards_2017, papernot_transferability_2016}. An adversarial example is a
deliberately crafted sample to fool the model's output by adding small but carefully
structured perturbations to a given input, which may or may not be perceptible to humans
(see Figure \ref{fig:adversarial-example}).

This questions the potential use of models in safety-critical systems and introduces
challenges that complicate regulatory approval for AI-based systems. Evaluating
adversarial robustness refers to the evaluation of the system's resilience and whether the
amount of perturbation needed to create the adversarial examples is sufficiently large,
either so that it becomes perceptible to humans, or so that it reasonably justifies the
model's error. As an initial step towards ensuring robustness, it is crucial to verify
that the adversarial examples identified involve a level of perturbation that makes the
model misclassification understandable. When trying to address adversarial robustness,
two major strategies have emerged from the community: (1) empirical defenses that
consist of architectural changes or training the model with adversarial examples,
to empirically improve the robustness against adversarial attacks
\cite{madry2017towards, dong2022adversarially}, (2) certified defenses that provide formal guarantees of the
model's robustness against input perturbations. A certificate $c$ is usually formally
defined for a given input $x$ as:

$$
\forall \delta, \quad \|\delta\| < c \implies f(x) = f(x + \delta)
$$
where $f(x)$ refers to the model's output.

Certified defenses are therefore a relevant way to assess the robustness of a model in
a safety-critical environment. To achieve that, several approaches are being developed
such as randomized smoothing \cite{cohen2019certified} providing probabilistic
certificates and formal verification methods \cite{katz2017reluplex, wang2021beta}.
Another approach is Lipschitz neural networks \cite{anil2019sorting} that are showing
promising results by controlling the Lipschitz constant of the model (see
Definition~\ref{def:lipschitz}).

\begin{definition}[Lipschitz Continuity]\label{def:lipschitz} A function \( f :
\mathbb{R}^n \rightarrow \mathbb{R}^m \) is said to be \textbf{Lipschitz continuous} if
there exists a constant \( K \geq 0 \) such that for all \( x_1, x_2 \in \mathbb{R}^n
\),
\begin{equation}
    \|f(x_1) - f(x_2)\| \leq K \|x_1 - x_2\|.
    \label{eq:lipschitz}
\end{equation}
The smallest such constant \( K \) is called the \textbf{Lipschitz constant} of \( f \),
denoted \( \operatorname{Lip}(f) \).
\end{definition}

To build Lipschitz neural networks, several methods can be used
\cite{miyato2018spectral,anil2019sorting, serrurier2021achieving}.  In practice, one can use libraries like
DEEL-Lip \cite{serrurier2021achieving} to
build such networks. Bounding the Lipschitz constant of neural networks has
been shown to improve robustness against adversarial attacks, generalization,
and interpretability \cite{hein2017formal, sokolic2017robust,
serrurier2021achieving, tsipras2018there}.

Combining this approach with adversarial attacks leads to both a lower bound and an
upper bound of the true robustness of the model on a given input. These certificates can
be used to certify a prediction and, when the bound is too low, trigger fallback
mechanisms, use more computationally expensive methods, such as formal methods, or human
intervention to ensure safe operation.

\section{Explainability}

\subsection{Definition and examples}

Explainability is defined in the ISO22989 (2022) \cite{ISO22989} as "the property of an
\ac{AI} system to express important factors influencing the \ac{AI} system results in a
way that humans can understand". This definition is human-centric, underscoring the
importance of making \ac{AI} systems transparent to all stakeholders and not only the
engineers developing it. Explainable \ac{AI} algorithms are already used for
production-grade models to provide insights for the data scientists or the business
experts into what the model focused on, and to support debugging and validation tasks.
Global \ac{xAI} explain how a model behaves in general and local \ac{xAI} explain the
reasons behind a specific prediction.

\begin{figure}[H]
    \centering
    \begin{subfigure} {0.6\textwidth}
        \centering
        \includegraphics[width=\linewidth]{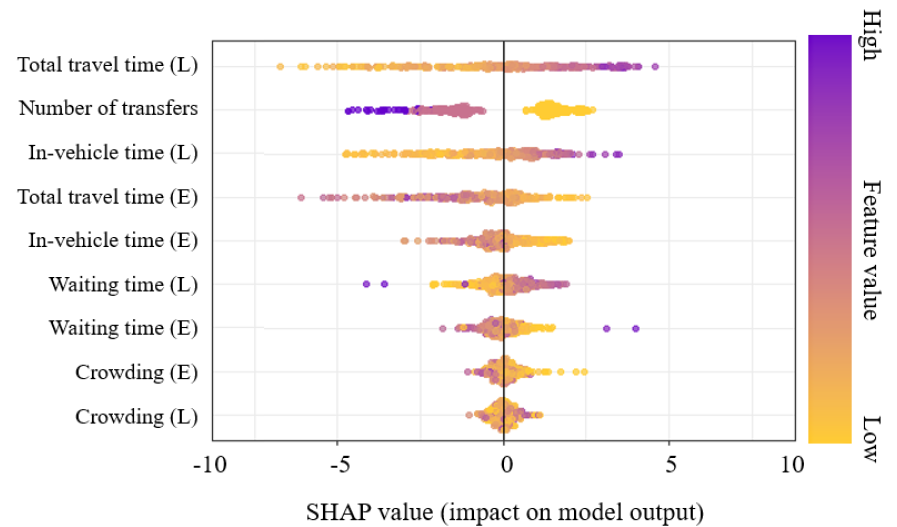}
        \caption{Global explainability of an express train classifier
        \cite{lee2021estimating}}
    \end{subfigure}
    \begin{subfigure} {0.3\textwidth}
        \centering
        \includegraphics[width=\linewidth]{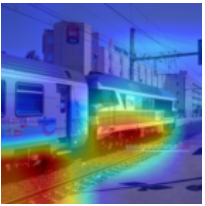}
        \caption{Local explainability of a classified train \cite{stalder2022what}}
    \end{subfigure}
    \caption{Global and local explainability}
    \label{fig:global_local_xai}
\end{figure}

Lee et al. (\citeyear{lee2021estimating}) \cite{lee2021estimating} present an example of
global interpretability of a model predicting the preference between using an express or
a local train from 9 features, using SHAP (\citeyear{TreeSHAP}) \cite{TreeSHAP}, a
feature attribution algorithm. Figure \ref{fig:global_local_xai} (a) shows the 9
features, ranked by importance. Each point represents one sample on which the model does
a prediction and gets a local interpretability with Shapley values
(\citeyear{shapley1953value}) \cite{shapley1953value} expressing the feature´s
contribution, as detailed in Figure \ref{fig:global_local_xai}. The local
interpretability values of all samples are plot all together. The scale on the right
shows the point´s feature values: purple is a high value of the feature, yellow is low.
The X-axis represents the impact on model output, negative or positive. In that case, a
negative value lowers the output probability of taking an express train, while a
positive value raises it. Engineers could draw conclusions from XAI, such as: the longer
the trip, the more likely a passenger is to take an express train according to the
model.

Figure \ref{fig:global_local_xai} (b) is an example of local explainability, of an image
classified as a train from Stalder et al. (\citeyear{stalder2022what})
\cite{stalder2022what}. GradCAM \cite{selvaraju2017grad}, a post-hoc attribution
algorithm, shows the pixel the algorithm focused on to classify. The more red the pixel
is, the more important it is for the computer vision classifier (in our example, the
algorithm focused on the boggie of the train). Local explainability methods are used in
the railway industry on production-grade algorithms, such as the detection and
recognition of wayside signals by Staino et al. (\citeyear{Staino2022})
\cite{Staino2022} or the detection and classification of wheel tread defects based
pictures by Trilla et al. (\citeyear{Trilla2021}) \cite{Trilla2021}. In railway systems,
apart from being a necessity to understand complex models, explainability is leveraged
for new use cases such as root cause analysis for incident investigation.

Both global and local explanations are necessary to provide engineers and users insights
on the system they are building.

\subsection{Methods and limitations}

Many state of the art publications list the methods for explainable \ac{AI} very well,
notably Molnar (\citeyear{Molnar}) \cite{Molnar} providing a deep review of algorithms
for tabular Machine Learning, the unified comprehensive review of Minh et all
(\citeyear{minh2022explainable}) \cite{minh2022explainable} and more recently Enemona et
al. (\citeyear{mathew2025recent}) \cite{mathew2025recent} on the emerging techniques in
2025.

Minh et all \cite{minh2022explainable} describe three groups of methods : (i)
pre-modeling explainability to ensure transparency in data preparation and feature
selection (ii) interpretable model, which are transparent by design (such as decision
trees, linear models, and rule-based systems) and (iii) post-modeling explainability,
refering to methods applied after a model has been trained, particularly for complex or
black-box models (for example SHAP for tabular data or Saliency maps for Computer
Vision). Fel et all \cite{fel2022xplique} gathered the most important methods of the
state of the art in a single package: Xplique, which facilitates the adoption. By
providing a consistent interface with XAI algorithms, the unified framework promotes
reproducibility and comparability across studies. Both SHAP and Gradcam, used for the
example of  Figure \ref{fig:global_local_xai}, are accessible.

However, we would like to highlight that even if the research on XAI algorithms seems
quite avanced, there are still important gaps before XAI becomes an enabler for
implementation in safety-related areas. The research community is more focused on
computer vision use cases than on tabular or time series applications, which are highly
important in the railway industry. The current state of the art presents hypothesis that
must be understood and tracked if explaination becomes a safety or a regulation
requirement. Indeed, post-modeling explainability algorithms are a representation of the
model´s reasoning, not the reality. For instance, one could expect the features
presented in Figure \ref{fig:global_local_xai} (a) to be quite correlated... and
computing explainations with SHAP when features are correlated is likely to provide an
interpretation not representing well the model´s decision making process, as widely
covered by the literature and by \cite{Molnar}. So mathematical hypothesis should be
tracked and verified if one is to provide trust on a model´s explainability.

The only safe way to ensure an explanation given is trustworthy as of now is to use a
model interpretable by design, which is what is currently done in regulated areas where
models take decision (ex: an expert model for interlocking presented by Klein et al.
(\citeyear{KLEIN1991499}) \cite{KLEIN1991499})). By using simple models such as linear
regression or decision trees, engineers and auditors can be sure to fully understand and
trust what´s behind the model decision, which is paramount for safety-related projects.
Another relevant research path is causality, which aims to identify a direct link
between an input and the model´s output. By understanding cause-and-effect relationships
in data rather than just identifying correlations, one can get back to the true reason
behind an artificial intelligence reasoning. However, this research domain is at its
early stage and will need more maturity before being fully integrated in production
products.

\section{Operational Design Domain}
As per DIN DKE SPEC 99004 \cite{DIN99004} (\citeyear{DIN99004}), the \acf{ODD} in Rail
is defined as an operational domain in which a given \ac{AI} system, or feature thereof,
is specifically designed to function. By clearly defining the boundaries, \ac{ODD} help
prevent \ac{AI} systems from operating outside their safe zones, reducing risk and
improving predictability. Confiance.AI provided a simple example of \ac{ODD} for their
welding quality use case \cite{welding2025}, where a model is used to detect welding
defects on images. Domain experts defined the acceptable conditions for image
acquisition: Image brightness, rotation angles between -30° and +30°, translation of the
piece in the image up to 5 millimeters.

\begin{figure}[ht]
    \centering
    \includegraphics[width=0.75\linewidth]{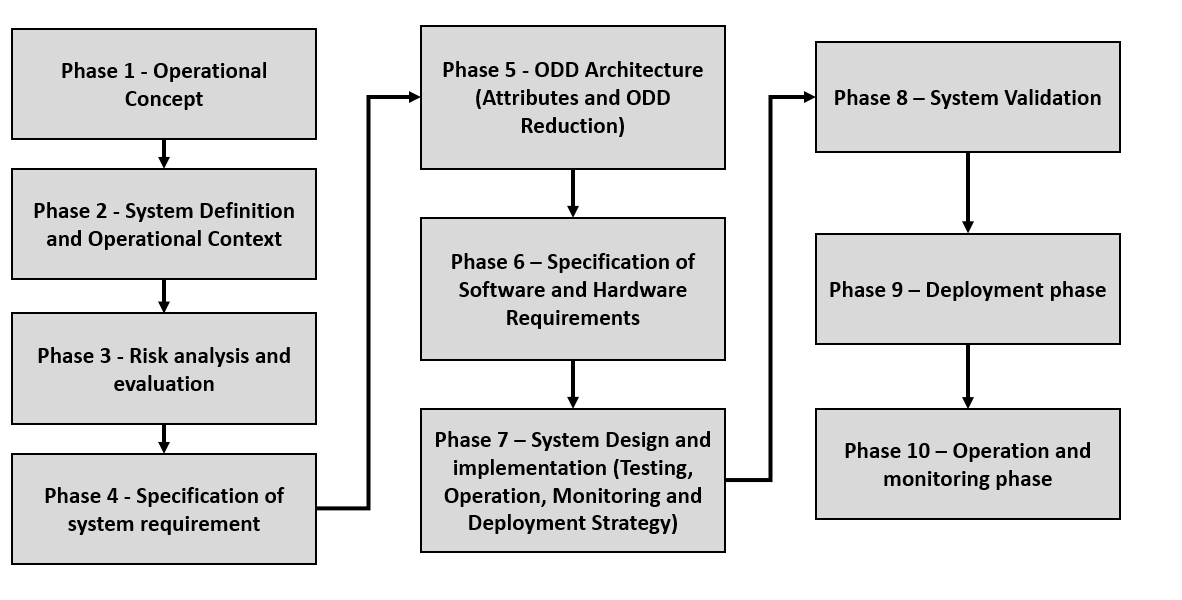}
    \caption{ODD Steps as per DIN DKE SPEC 99004 \cite{DIN99004}}
    \label{fig:lifecycle_ODD}
\end{figure}

The DIN DKE SPEC 99004 \cite{DIN99004} details the steps to follow for the ODD´s
lifecycle. We propose to resume this lifecycle in Figure \ref{fig:lifecycle_ODD}.

The research on ODD has been primarily driven by the Autonomous Driving field, which
happens to use Artificial Intelligence in some specific cases. We believe that what was
done for Autonomous Driving could be applied to all A.I use cases, which would bring
more trust in all production algorithms. The ODD concepts are already well defined and
industrialized for autonomous driving, and all AI applications would benefit from the
standardized requirements already guiding the development of autonomous vehicles.
Indeed, when reading the 10 purposes of an ODD presented in \cite{DIN99004}, one can
identify attributes that are desirable for any AI system. For instance: "ODD limit
detection: By defining an ODD one can develop algorithms to determine which sensing
capabilities or data input are needed to detect when the AI System is inside and outside
expected operating conditions." That statement would ring a bell to any data scientist
who have been confronted to the issue of data drift.

Data drift is defined as a significant change to the data distribution compared to the
data used for training. The model is then operating over a domain it has not been
properly evaluated on, which may decrease the trust one could have over the results.
Many methods exist for data drift detection in an industrial context, as detailed by
Simonetto et al. \cite{INDUSTRIAL_DRIFT} (\citeyear{INDUSTRIAL_DRIFT}), ranging from
statistical tests to uncertainty evaluation. So detecting when the input data is
different from the boundaries defined in the \ac{ODD} and the training data is feasible.
The issue is more on the actions taken after a drift identification: if a model
retraining is required, then change management of the \ac{ODD} is also needed, which is
costly and long. Hence the need for the standardization and industrialization of ODD, to
save engineering and validation time for changes.


Overall, \ac{ODD} is an important lever of trust for \ac{AI} solutions. The strong added
value of \ac{ODD} highlights the key need of system engineers when designing \ac{AI}
solutions, and not only data scientists focusing on creating the best algorithms. 

\section{A System Approach to Trustworthy AI}

Rather than treating robustness, ODD, and explainability as isolated requirements, we
advocate for a systemic approach where these elements interact continuously. Those three
pillars shall be integrated in operations to continuously monitor \ac{AI} performance
and boundary conditions, and identify when a model shall be updated or not used at all.
System definitions are already defined in detail in the work of Confiance.AI by Mattioli
et al. in several publications \cite{mattioli2024overview, quinterohal05074120,
mattiolihal04813492} (\citeyear{mattiolihal04813492}).

\begin{figure}[ht]
    \centering
    \includegraphics[width=0.95\linewidth]{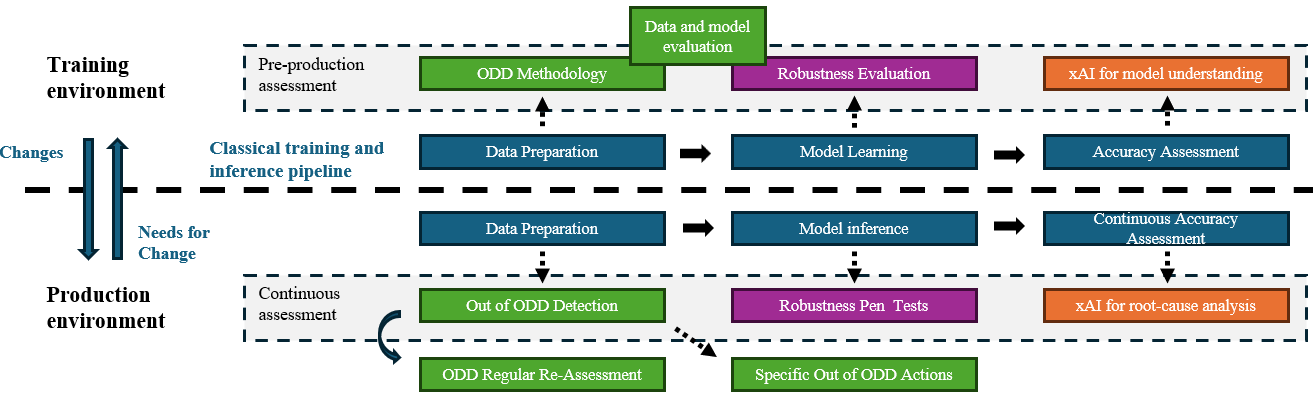}
    \caption{System approach using Trustworthy \ac{AI} technologies}
    \label{fig:system_view}
\end{figure}

We propose in Figure \ref{fig:system_view} an approach where those three technological
bricks are integrated into a classical Machine Learning flow (in blue): The \ac{ODD} is
defined after having acquired the right data and the robustness tests are performed on
the model continuously. In production, the ODD technologies are used to identify needs
for changes in the ODD (in green) and take mitigation actions for specific samples. The
explainability bricks (in orange) are used by engineers to understand the model,
globally in a training environment, then locally to understand each prediction of the
production environment.

To do so, a safe MLOps (Machine Learning Operations) environment is necessary, as
proposed by Zeller et al. \cite{zeller2023towards} (\citeyear{zeller2023towards}), with
a software infrastructure able to continuously assess and validate the trustworthiness
of an \ac{AI} in production without a human intervention. This will faster development
by simplifying the engineering workload of monitoring models in production and increase
cross-functional collaboration between \ac{AI} developers and safety engineers. Such a
system fosters trust not only from regulators but also from operators and the public,
accelerating \ac{AI} adoption in mission-critical railway applications. A system
approach is imperative if one wants to deploy \ac{AI} products in critical environments.

\section{Conclusion}

By focusing on robustness, \ac{ODD} and \ac{xAI} —and integrating them into a systemic
framework— we can pave the way for safe, compliant, and widely accepted \ac{AI} systems.
Future work should focus on the development of certification frameworks and standards
that establish the foundation to clear and exhaustive requirements for safety-critical
applications in railways. Looking for algorithm compliance is not possible without the
necessary norms to frame the use of AI. Railways would benefit from cross-industry
collaboration to leverage their best practices and standards. Several sectors are more
advanced, such as the aerospace and automotive where engineers are already implementing
\ac{AI} models in safety-critical systems, supported by explicit requirements. This
publication does not exhaustively treat all key attributes needed for a trustworthy AI,
which should operate ethically and safely, within a cyber-secured environment and be
supported by a strong data management system that correctly stores accurate data. Only
by being extremely precise over the complete life cycle of \ac{AI} will we be able to
deploy \ac{AI} in railway safety-critical environments.

\printbibliography

\end{document}